\documentclass[a4paper]{article}
\usepackage{dx}  
\usepackage{times}  
\usepackage{helvet} 
\usepackage{courier}  
\usepackage[hyphens]{url}  
\usepackage{graphicx} 
\usepackage{graphicx}  
\usepackage{latexsym}

\usepackage{subfig}

\usepackage{abraces}

\usepackage{soul}
\usepackage[utf8]{inputenc}
\usepackage{amsmath}
\usepackage{booktabs}
\usepackage{amsthm}
\usepackage{amsfonts}
\usepackage[T1]{fontenc}

\usepackage{quoting}

\usepackage[multiple]{footmisc}

\usepackage{fancybox}
\usepackage{amsfonts}

\usepackage{environ}
\usepackage{mathnotation}
\usepackage{multirow}

\usepackage{color}

\usepackage{xspace}
\usepackage{dashrule}
\usepackage[table,xcdraw,dvipsnames]{xcolor}
\usepackage[inline]{enumitem}

\usepackage{footmisc}

\newcommand{\dpi}{\mathit{dpi}}

\newcommand{\dt}{\mathcal{D}^*}
\newcommand{\md}{\mathcal{D}}
\newcommand{\mD}{\mathbf{D}}
\newcommand{\mC}{\mathbf{C}}

\newcommand{\ld}{\mathit{ld}}
\newcommand{\mo}{\mathcal{K}}
\newcommand{\mb}{\mathcal{B}}
\newcommand{\tax}{\mathit{ax}}
\newcommand{\Tp}{\mathit{P}}
\newcommand{\Tn}{\mathit{N}}
\newcommand{\tp}{\mathit{p}}
\newcommand{\tn}{\mathit{n}}
\newcommand{\mc}{\mathcal{C}}

\newcommand{\pr}{\mathit{pr}}
\newcommand{\bnd}{\mathit{bound}}
\newcommand{\node}{\mathsf{n}}
\newcommand{\valid}{\mathit{valid}}
\newcommand{\closed}{\mathit{closed}}
\newcommand{\childnodes}{\mathsf{Child\_Nodes}}

\usepackage{algorithm}
\usepackage[noend]{algpseudocode}
\algrenewcommand\algorithmicrequire{\textbf{Input:}}
\algrenewcommand\algorithmicensure{\textbf{Output:}}
\algrenewcommand\alglinenumber[1]{\tiny #1:} 
\algnewcommand{\IfThen}[2]{
	\State \algorithmicif\ #1\ \algorithmicthen\ #2
}

\usepackage[all]{xy}
\usepackage{tikz}

\newcounter{examplecounter}
\newenvironment{example}{
	\refstepcounter{examplecounter}%
	
	\vspace{7pt}
	\noindent\textbf{Example \arabic{examplecounter}}%
	\quad
}{
	
	\vspace{7pt}
}

\newtheorem{thm}{Theorem}

\title{Sound, Complete, Linear-Space, Best-First Diagnosis Search\thanks{This work was accepted for presentation at the \emph{31st International Workshop on Principles of Diagnosis (DX-2020)}. Versions of this paper were published in the \emph{Proceedings of the International Symposium on Combinatorial Search (SoCS-2021)} \protect\cite{rodler2021socs} as well as in the \emph{Artificial Intelligence Journal (AIJ)} \protect\cite{rodler2022memorylimited}. The latter version includes i.a.\ a more thorough discussion of related work, complexity analyses, all proofs, an extended evaluation section, more examples, and a more comprehensive description of the algorithm.}}
\author{
	Patrick Rodler\textsuperscript{\rm 1}\\ 
\textsuperscript{\rm 1}University of Klagenfurt, Austria, \\ patrick.rodler@aau.at 
}
 
\begin{document}

\maketitle
%

\begin{abstract}
Various model-based diagnosis scenarios
require the computation of the most preferred fault explanations. Existing algorithms that are sound (i.e., output \emph{only} actual fault explanations) and complete (i.e., can return \emph{all} explanations), however, require exponential space to achieve this task. As a remedy, to enable successful diagnosis 
on memory-restricted devices and for memory-inten-sive problem cases, we propose RBF-HS, a diagnostic search 
based on Korf's well-known RBFS algorithm. RBF-HS can enumerate an arbitrary fixed number of fault explanations in best-first order within linear space bounds, without sacrificing the desirable soundness or completeness properties. 
Evaluations using real-world diagnosis cases show that RBF-HS, when used to compute minimum-cardinality fault explanations, in most cases saves substantial space (up to 98\,\%) while requiring only reasonably more or even less time than Reiter's HS-Tree, 
a commonly used and
as generally applicable sound, complete and best-first diagnosis search.

\end{abstract}

\section{Introduction}
\label{sec:intro}
In model-based diagnosis, heuristic search techniques have proven to be a powerful tool for the computation of parsimonious fault explanations (\emph{minimal diagnoses}) \cite{jannach2016parallel,wotawa2001variant,lin2003computation,greiner1989correction,Shchekotykhin2014,rodler2018statichs,rodler2020ecai,feldman2006two}. Due to the 
(NP-)hardness of the diagnosis computation problem,\footnote{Given a non-empty set of minimal diagnoses, computing another minimal diagnosis is NP-hard, even if theorem proving is in P \cite{Bylander1991}.} it is in most applications infeasible to determine all (minimal) diagnoses. 

For this reason---besides the often stipulated guarantee that only minimal diagnoses are generated (\emph{soundness}) and no minimal diagnosis is missed (\emph{completeness})---the focus of 
search techniques is usually laid on the \emph{best} minimal diagnoses, e.g., the most probable or minimum-cardinality ones.
In particular, the set of the best 
diagnoses appears to be more appropriate than just \emph{any} sample of diagnoses, i.a.,   
\begin{itemize}[noitemsep]
	\item if all components have a very low failure probability 
	(actual diagnosis 
	among 
	minimum-cardinality diagnoses),
	as is the case for many physical devices,
	\item if the given probabilistic information is trustworthy and well-founded 
	(actual diagnosis among most probable diagnoses),
	like
	in 
	projects with a long history of bug-fixes, or  
	\item in sequential diagnosis\footnote{\emph{Sequential diagnosis} \cite{dekleer1987} aims at the gradual elimination of spurious 
		diagnoses
		through the suggestion of informative system measurements, until some stop criterion is met, e.g., only one diagnosis remains. We say that sequential diagnosis relies on \emph{early termination} if it might already stop although multiple diagnoses are still possible, e.g., if the probability of some diagnosis exceeds some predefined threshold.} scenarios where an early termination appears to be reasonable only if the best remaining solution is always known.
\end{itemize}
%

%
Like for many algorithms, there is a trade-off between time and space complexity for diagnosis search methods.
 Among the factors time and space, the latter
can be the more critical criterion.
Because, if the memory consumption of an algorithm exceeds the amount of available memory, the problem becomes intractable, whereas, with a higher time demand, an algorithm does still work in principle and will deliver a solution (albeit with a potentially longer waiting time). In model-based diagnosis, there is a range of scenarios which \emph{(a)}~pose substantial memory requirements on diagnostic search methods or \emph{(b)}~suffer from too little memory. 
One example for (a) are problems involving high-cardinality diagnoses, e.g., when two systems are integrated and a multitude of errors emerge at once
\cite{Shchekotykhin2014,meilicke2011thesis}. Manifestations of (b) are frequently found in today's era of the Internet of Things (IoT), distributed or autonomous systems, and ubiquitous computing, where low-end microprocessors, often with only a small amount of RAM, are built into almost any device. Whenever such devices should perform (self-)diagnosing actions \cite{klein1999evaluating,williams1996model}, memory-aware diagnosis algorithms are a must \cite{williams2007conflict,zoeteweij2008automated}. 
Existing (sound and complete) best-first diagnosis search methods require an exponential amount of memory in that all paths in the search tree must be stored in order to guarantee that the best one is expanded in each iteration. Hence, they often disqualify for scenarios like (a) and (b) above. 
As a remedy, 
we have devised a diagnosis search called \emph{Recursive Best-First Hitting Set Search (RBF-HS)}, which is based on 
Korf's well-known \emph{Recursive Best-First Search (RBFS)} algorithm \cite{korf1992linear}.
 RBF-HS features all the desirable properties of diagnostic searches, i.e., \emph{soundness}, \emph{completeness}, and the \emph{best-first property}, and is able to return an arbitrary fixed number of the best existing solutions within \emph{linear memory} bounds. Moreover, RBF-HS is \emph{generally applicable} regardless of the (monotonic) system description language or the particular 
 logical theorem prover used.


In extensive evaluations 
on minimum-cardinality diagnosis computation tasks over 
real-world diagnosis cases, 
we compare the performance of RBF-HS with that of Reiter's HS-Tree, a state-of-the-art sound, complete and best-first diagnosis search that is as generally applicable as RBF-HS.
The results evince that 
RBF-HS
 is comparable with HS-Tree in terms of runtime
even though the latter requires significantly (up to orders of magnitude) more memory.

\section{Preliminaries}
\label{sec:basics}
We first briefly characterize 
MBD concepts used throughout this work, based on the framework of \cite{Shchekotykhin2012,Rodler2015phd} which is (slightly) more general \cite{rodler17dx_reducing} than Reiter's theory \cite{Reiter87}.\footnote{The main reason for using this more general framework is its ability to handle negative measurements (things that must \emph{not} be true for the diagnosed system)
which are helpful, e.g., for diagnosing knowledge bases \cite{DBLP:journals/ai/FelfernigFJS04,Shchekotykhin2012}.} 

\noindent\textbf{Diagnosis Problem.} 
We assume that the diagnosed system, consisting of a set of components $\setof{c_1,\dots,c_k}$, is described by a finite set of logical sentences $\mo \cup \mb$, where $\mo$ (possibly faulty sentences) includes knowledge about the behavior of the system components, and $\mb$ (correct background knowledge) comprises any additional available system knowledge and system observations. More precisely, there is a one-to-one relationship between sentences $\tax_i \in \mo$ and components $c_i$, where $\tax_i$ describes the nominal behavior of $c_i$ (\emph{weak fault model}). E.g., if $c_i$ is an AND-gate in a circuit, then $\tax_i := out(c_i) = and(in1(c_i),in2(c_i))$; $\mb$ in this case might contain sentences stating, e.g., which components are connected by wires, or observed circuit outputs. 
The inclusion of a sentence $\tax_i$ in $\mo$ corresponds to the (implicit) assumption that $c_i$ is healthy. Evidence about the system behavior is captured by sets of positive ($\Tp$) and negative ($\Tn$) measurements \cite{dekleer1987,Reiter87,DBLP:journals/ai/FelfernigFJS04}. Each measurement is a logical sentence; positive ones $\tp\in\Tp$ must be true and negative ones $\tn\in\Tn$ must not be true. The former can be, depending on the context, e.g., observations about the system, probes or required system properties. 
The latter model properties that must not hold for the system, e.g., if $\mo$ is a biological knowledge base to be debugged, a negative test case might be $\forall X \mathit{bird}(X) \to \mathit{flies}(X)$ (``every bird flies'').
We call $\tuple{\mo,\mb,\Tp,\Tn}$ a \emph{diagnosis problem instance (DPI)}. 

\begin{example} \hspace{-1em}\emph{(Diagnosis Problem)}\quad\label{ex:dpi}
	Tab.~\ref{tab:example_DPI} depicts a DPI stated in propositional logic. 
	The ``system'' (which is the knowledge base itself in this case) comprises five ``components'' $c_1, \dots,c_5$, and the ``nominal behavior'' of $c_i$ is given by the respective axiom $\tax_i \in \mo$. There is neither any background knowledge ($\mb = \emptyset$) nor any positive measurements ($\Tp=\emptyset$) available from the start. But, there is one negative measurement (i.e., $\Tn = \setof{\lnot A}$), which postulates that $\lnot A$ must \emph{not} be an entailment of the correct system (knowledge base). Note, however, that $\mo$ (i.e., the assumption that all ``components'' work nominally) in this case does entail $\lnot A$ (e.g., due to the axioms $\tax_1,\tax_2$) and therefore some axiom in $\mo$ must be faulty (i.e., some ``component'' is not healthy).\qed
\end{example}

\begin{table}
	\centering
	\scriptsize
	\renewcommand{\arraystretch}{1}
	\begin{tabular}{@{}ccc@{}}
		\toprule
		\multicolumn{1}{ c  }{\multirow{2}{*}{$\mo\;=$} } & \multicolumn{2}{ l  }{$\{ \tax_1: A \to \lnot B$ \;\; $\tax_2: A \to B$ \;\; $\tax_3: A \to \lnot C$} \\
		& \multicolumn{2}{ l  }{$\phantom{\{} \tax_4: B \to C$ \;\;\,\hspace{4pt} $\tax_5: A \to B \lor C \qquad\qquad\,\quad\;\}$} \\
		\cmidrule{1-3}
		\multicolumn{3}{ l  }{$\mb =\emptyset \quad\qquad\qquad\qquad \Tp=\emptyset \quad\qquad\qquad\qquad \Tn=\setof{\lnot A}$} \\
		\bottomrule
	\end{tabular}
	\caption{\small Example DPI stated in propositional logic.}
	\label{tab:example_DPI}
\end{table}

\noindent\textbf{Diagnoses.} 
If the system description along with the positive measurements (under the 
assumption $\mo$ that all components are healthy) is inconsistent, i.e., $\mo \cup \mb \cup \Tp \models \bot$, or some negative measurement is entailed, i.e., $\mo \cup \mb \cup \Tp \models \tn$ for some $\tn \in \Tn$, some assumption(s) about the healthiness of components, i.e., some sentences in $\mo$, must be retracted. We call such a set of sentences $\md \subseteq \mo$ a \emph{diagnosis} for the DPI $\tuple{\mo,\mb,\Tp,\Tn}$ iff $(\mo \setminus \md) \cup \mb \cup \Tp \not\models x$ for all $x \in \Tn \cup \setof{\bot}$. We say that $\md$ is a \emph{minimal diagnosis} for $\dpi$ iff there is no diagnosis $\md' \subset \md$ for $\dpi$. The set of minimal diagnoses is representative of all diagnoses under the weak fault model \cite{Kleer1992}, i.e., 
the set of all diagnoses 
is equal to 
the set of all supersets of minimal diagnoses.
Thus, diagnosis approaches usually restrict their focus to only minimal diagnoses. 
We furthermore denote by $\dt$ the \emph{actual diagnosis} which pinpoints the actually faulty axioms, i.e., all elements of $\dt$ are in fact faulty and all elements of $\mo\setminus\dt$ are in fact correct.

\begin{example} \hspace{-1em}\emph{(Diagnoses)}\quad\label{ex:diagnoses}
	For our DPI in Tab.~\ref{tab:example_DPI} we have four minimal diagnoses, given by $\md_1:=[\tax_1,\tax_3]$, $\md_2:=[\tax_1,\tax_4]$, $\md_3:=[\tax_2,\tax_3]$, and $\md_4 := [\tax_2,\tax_5]$.
	For instance, $\md_1$ is a minimal diagnosis as $(\mo\setminus\md_1) \cup \mb\cup \Tp = \setof{\tax_2,\tax_4,\tax_5}$ is both consistent and does not entail the given negative measurement $\lnot A$.\qed
\end{example}

\noindent\textbf{Diagnosis Probability Model.} 
In case useful meta information is available that allows to assess the likeliness of failure for system components, the probability of diagnoses (of being the actual diagnosis) can be derived.
Specifically, given a function $\pr$ that maps each sentence (system component) $\tax \in \mo$ to its failure probability $\pr(\tax)\in (0,1)$, the probability $\pr(X)$ of a diagnosis candidate\footnote{Note, the probability (of being equal to the actual diagnosis) of some $X \subseteq \mo$ which is not a diagnosis is trivially zero. 
Still, it is reasonable to define the probability $\pr$ for such sets as well. The reason is that 
diagnosis searches
like the one discussed in this work grow diagnoses 
stepwise, starting from the empty set, and it can make a substantial difference (in terms of performance), which of those partial diagnoses are further explored when. To this end, the probabilities 
can provide a valuable guidance.} $X \subseteq \mo$ (under the common assumption of independent component failure) is computed as the probability that all sentences in $X$ are faulty, and all others are correct, i.e., $\pr(X) := \prod_{\tax \in X} \pr(\tax) \prod_{\tax \in \mo\setminus X} (1-\pr(\tax))$.

\begin{example} \hspace{-1em}\emph{(Diagnosis Probabilities)}\quad\label{ex:diag_probs}
	Let the component probabilities for the DPI in Tab.\ref{tab:example_DPI} be
	$\langle \pr(\tax_1),\dots$, $\pr(\tax_5)\rangle$ $=$ $\langle .1, .05, .1, .05, .15\rangle$. Then, we can compute the probabilities of all minimal diagnoses from Example~\ref{ex:diagnoses} as $\langle \pr(\md_1), \dots, \pr(\md_4)\rangle = \langle .0077,.0036,.0036,.0058\rangle$. For instance, $\pr(\md_1)$ is calculated as $.1*(1-.05)*.1*(1-.05)*(1-.15)$. The normalized diagnoses probabilities would then be $\langle .37,.175,.175,.28\rangle$. Note, this normalization makes sense if not all diagnoses, but only \emph{minimal} diagnoses are of interest, which is usually the case in model-based diagnosis applications for complexity reasons.\qed 
\end{example}

\noindent\textbf{Conflicts.} 
Useful for diagnosis computation 
is the notion of a conflict \cite{dekleer1987,Reiter87}.
A conflict is a set of healthiness assumptions for components $c_i$ that cannot all hold given the current knowledge about the system. More formally, $\mc \subseteq \mo$ is a \emph{conflict} for the DPI $\tuple{\mo,\mb,\Tp,\Tn}$ iff $\mc \cup \mb \cup \Tp \models x$ for some $x \in \Tn \cup \setof{\bot}$. We call $\mc$ a \emph{minimal conflict} for $\dpi$ iff there is no conflict $\mc' \subset \mc$ for $\dpi$.
%


\begin{example} \hspace{-1em}\emph{(Conflicts)}\quad\label{ex:conflicts}
	For our running example, $\dpi$, in Tab.~\ref{tab:example_DPI}, there are four minimal conflicts, given by $\mc_1 := \langle\tax_1,\tax_2\rangle$, $\mc_2 := \langle\tax_2,\tax_3,\tax_4\rangle$, $\mc_3 := \langle\tax_1,\tax_3,\tax_5\rangle$, and $\mc_4 := \langle\tax_3,\tax_4,\tax_5\rangle$.
	For instance, $\mc_4$, in CNF equal to $(\lnot A \lor \lnot C) \land (\lnot B \lor C) \land (\lnot A \lor B \lor C)$, is a conflict because adding the unit clause $(A)$ to this CNF yields a contradiction, which is why the negative test case $\lnot A$ is an entailment of $\mc_4$. The minimality of the conflict $\mc_4$ can be verified by rotationally removing from $\mc_4$ a single axiom at the time and controlling for each so obtained subset that this subset is consistent and does not entail $\lnot A$.\qed
\end{example}

\noindent\textbf{Relationship between Conflicts and Diagnoses.}
Conflicts and diagnoses are closely related in terms of a hitting set and a duality property \cite{Reiter87}: 
\begin{description}[font=\normalfont\em]
	\item[Hitting Set Property] A (minimal) diagnosis for $\dpi$ is a (minimal) hitting set of all minimal conflicts for $\dpi$. \\ ($X$ is a \emph{hitting set} of a collection of sets $\mathbf{S}$ iff 
	$X \subseteq \bigcup_{S_i \in \mathbf{S}} S_i$ and $X \cap S_i \neq \emptyset$ for all $S_i \in S$; a hitting set $X$ is \emph{minimal} iff there is no other hitting set $X'$ with $X' \subset X$)
	\item[Duality Property] Given a DPI $\dpi = \tuple{\mo,\mb,\Tp,\Tn}$, $X$ is a diagnosis (or: contains a minimal diagnosis) for $\dpi$ iff $\mo \setminus X$ is not a conflict (or: does not contain a minimal conflict) for $\dpi$.
\end{description}

\section{RBF-HS Algorithm}
\label{sec:algo}


\noindent\textbf{The Idea.}
Korf's heuristic search algorithm RBFS \cite{korf1992linear} provides the inspiration for RBF-HS. Historically, the main motivation that led to the engineering of RBFS was the problem that best-first searches by that time required exponential space. The goal of RBFS is to trade (more) time for (much less) space by means of a 
``\emph{(re)explore-current-best} \& \emph{backtrack} \& \emph{forget-most} \& \emph{remember-essential} \& \emph{update-cost}'' cycle. In this vein, RBFS 
works within linear-space bounds while maintaining completeness and  the best-first property.

\begin{algorithm}[h!]
	\scriptsize
	\caption{RBF-HS} \label{algo:RBF_HS}
	{\fontsize{7pt}{8pt}\selectfont
		\begin{algorithmic}[1]
			\Require 
			\textcolor{white}{.}
			tuple $\tuple{\dpi, \pr, \ld}$ comprising
			\begin{itemize}[noitemsep]
				\item a DPI $\dpi = \langle\mo,\mb,\Tp,\Tn\rangle$
				\item a probability measure $\pr$ that assigns a failure probability $\pr(\tax) \in (0,1)$ to each $\tax \in \mo$ (cf.\ Sec.~\ref{sec:basics}), where $\pr$ is cost-adjusted; \emph{note:} the cost function $f(\node) := \pr(\node)$ for all tree nodes $\node$
				\item the number $\ld$ of leading minimal diagnoses to be computed 
			\end{itemize}
			\Ensure 
			list $\mD$ where $\mD$ is the list of the $\ld$ (if existent) most probable (as per $\pr$) 
			minimal diagnoses wrt.\ $\dpi$, sorted by probability in descending order  
			
			\vspace{6pt}
			\Procedure{RBF-HS}{$\dpi, \pr, \ld$}
			\State $\mD \gets [\,], \; \mC \gets [\,]$
			\label{algoline:rbfhs:initialize_mD_mC_f}
			\State $\mc \gets \Call{FindMinConflict}{\dpi}$ \label{algoline:rbfhs:findMinConflict}
			\If{$\mc =\emptyset$}  \label{algoline:rbfhs:mc=emptyset}
			\State \Return $\mD$ \label{algoline:rbfhs:return_mD_1}	
			\EndIf
			\If{$\mc = \text{'no conflict'}$}  \label{algoline:rbfhs:mc='no_conflict'}
			\State \Return $[\emptyset]$ \label{algoline:rbfhs:return_mD_2}	
			\EndIf
			\State $\mC \gets \Call{add}{\mc,\mC}$  \label{algoline:rbfhs:add_mc_to_mC}	
			\State $\Call{RBF-HS'}{\emptyset,f(\emptyset),-\infty}$ \label{algoline:rbfhs:call_RBFHS'} \Comment{$\emptyset$ is the root node}
			\State \Return $\mD$	\label{algoline:rbfhs:return_mD_3}																															
			\EndProcedure
			\vspace{6pt}
			
			\Procedure{RBF-HS'}{$\mathsf{\node}, F(\mathsf{\node}), \bnd$} \label{algoline:rbfhs':procedure_RBF-HS'}
			\State $L \gets \Call{label}{\node}$ \label{algoline:rbfhs':label}
			\If{$L = \closed$} \label{algoline:rbfhs':if_nonmin}
			\State \Return $-\infty$ \label{algoline:rbfhs':return_after_closed}
			\EndIf
			\If{$L = \valid$} \label{algoline:rbfhs':if_valid}
			\State $\mD \gets \Call{add}{\node,\mD}$ \label{algoline:rbfhs:add_node_to_mD}  \Comment{new minimal diagnosis found}
			\If{$|\mD| \geq \ld$}    \label{algoline:rbfhs':if_mD_geq_ld}
			\State \textbf{exit procedure} \label{algoline:rbfhs':exit_procedure}
			\EndIf
			\State \Return $-\infty$   \label{algoline:rbfhs':return_after_valid}
			\EndIf
			\State $\childnodes \gets \Call{expand}{\node,L}$ \label{algoline:rbfhs':expand}
			\For{$\node_i \in \childnodes$} \label{algoline:rbfhs':for_node_in_childnodes}
			\If{$f(\node) > F(\node)$} \label{algoline:rbfhs':if_f(n)>F(n)} \Comment{if $\true$, $\node$ was already expanded before}
			\State $F(\node_i) \gets \min(F(\node),f(\node_i))$ \label{algoline:rbfhs':F(n_i)_gets_min}
			\Else
			\State $F(\node_i) \gets f(\node_i)$ \label{algoline:rbfhs':F(n_i)_gets_f(n_i)}
			\EndIf
			\EndFor 
			\If{$|\childnodes|=1$} \label{algoline:rbfhs':if_|childnodes|=1} \Comment{add dummy node $\node_d$ with $F(\node_d) = -\infty$}
			\State $\childnodes \gets \Call{addDummyNode}{\childnodes}$
			\EndIf
			\State $\childnodes \gets \Call{sortDecreasingByF}{\childnodes}$ \label{algoline:rbfhs':sortDecreasingByF}
			\State $\node_1 \gets \Call{getAndDeleteFirstNode}{\childnodes}$ \label{algoline:rbfhs':getBestChild_1} \Comment{$\node_1\dots$best child}
			\State $\node_2 \gets \Call{getFirstNode}{\childnodes}$ \label{algoline:rbfhs':getSecondBestChild_1} \Comment{$\node_2\dots$2nd-best child}
			\While{$F(\node_1) \geq \bnd \;\land\; F(\node_1) > -\infty$} \label{algoline:rbfhs':while}
			\State $F(\node_1) \gets \Call{RBF-HS'}{\node_1,F(\node_1),\max(\bnd,F(\node_2))}$ \label{algoline:rbfhs':recursive_call}
			\State $\childnodes \gets \Call{insertSortedByF}{\node_1, \childnodes}$ \label{algoline:rbfhs':insertSortedByF}
			\State $\node_1 \gets \Call{getAndDeleteFirstNode}{\childnodes}$ \label{algoline:rbfhs':getBestChild_2} \Comment{$\node_1\dots$best child}
			\State $\node_2 \gets \Call{getFirstNode}{\childnodes}$ \label{algoline:rbfhs':getSecondBestChild_2} \Comment{$\node_2\dots$2nd-best child}
			\EndWhile
			\State \Return $F(\node_1)$ \label{algoline:rbfhs':return_F(n)}
			\EndProcedure
			\vspace{6pt}
			
			\Procedure{\textsc{label}}{$\node$} 
			\For{$\node_i \in \mD$}\label{algoline:label:non-min_crit_start}
			\If{$\node \supseteq \node_i$}  \Comment{goal test, part 1 (is $\node$ non-minimal?)} \label{algoline:label:if_n_supseteq_n_i}  
			\State \Return $\closed$ \Comment{$\node$ is a non-minimal diagnosis}
			\EndIf
			\EndFor\label{algoline:label:non-min_crit_end}
			\For{$\mc \in \mC$}\label{algoline:label:reuse_start}
			\If{$\mc \cap \node = \emptyset$}\label{algoline:label:if_C_cap_node=emptyset}  \Comment{cheap non-goal test (is $\node$ not a diagnosis?)}   
			\State \Return $\mc$\label{algoline:label:return_C} \Comment{$\node$ is not a diagnosis; reuse $\mc$ to label $\node$}
			\EndIf 
			\EndFor\label{algoline:label:reuse_end}
			\State $L\gets \Call{findMinConflict}{\langle\mo\setminus\node,\mb,\Tp,\Tn\rangle}$\label{algoline:label:findMinConflict}  
			\If{$L$ = \text{'no conflict'}}	 \Comment{goal test, part 2 (is $\node$ diagnosis?)} \label{algoline:label:'no_conflict'}					
			\State \Return $\valid$\label{algoline:label:return_valid} \Comment{$\node$ is a minimal diagnosis}
			\Else	\Comment{$\node$ is not a diagnosis}
			\State $\mC \gets \Call{add}{L,\mC}$\label{algoline:label:add_new_cs}  \Comment{$L$ is a \emph{new} minimal conflict ($\notin \mC$)}
			\State \Return $L$\label{algoline:label:return_new_cs}
			\EndIf
			\EndProcedure
			
			\vspace{6pt}
			
			\Procedure{\textsc{expand}}{$\node, \mc$}
			\State $\mathsf{Succ\_Nodes} \gets [\,]$
			\For{$e \in \mc$} \label{algoline:expand:for_loop}
			\State $\mathsf{Succ\_Nodes} \gets \Call{add}{\node\cup\{e\},\mathsf{Succ\_Nodes}}$ \label{algoline:expand:add_successor_node}
			\EndFor
			\State \Return $\mathsf{Succ\_Nodes}$
			\EndProcedure	
		\end{algorithmic}
	}
	\normalsize
\end{algorithm}

\noindent\textbf{Notation.}
RBF-HS is depicted by Alg.~\ref{algo:RBF_HS}. It deals with nodes, where each node $\node$ is a subset of or equal to a diagnosis and corresponds to a set of edge labels along one branch from the root of the constructed hitting set tree (cf.\ \cite{Reiter87}). Nodes can be unlabeled (initially, after being generated) or labeled by either $\valid$ (node is a minimal diagnosis), $\closed$ (node is a non-minimal or already computed minimal diagnosis), or by a minimal conflict (node is a subset of a diagnosis). There are two functions, $f$ and $F$, that assign a cost to each node, where $f$ defines initial costs ($f:=\pr$) and remains constant throughout the execution of RBF-HS, and $F$ specifies backed-up (or: learned) costs and is subject to change while RBF-HS runs. 

\noindent\textbf{Inputs and Output.}
 RBF-HS accepts 
a DPI $\dpi = \tuple{\mo,\mb,\Tp,\Tn}$,
a probability measure $\pr$ (see Sec.~\ref{sec:basics}), and 
a stipulated number $\ld$ of minimal diagnoses to be returned as input arguments. 
It outputs the $\ld$ (if existent) most probable (wrt.\ $\pr$) minimal diagnoses 
for $\dpi$. 

Note, for correctness reasons \cite{Rodler2015phd}, the probability model $\pr$ must be \emph{cost-adjusted}, i.e., $\pr(\tax) < 0.5$ for all $\tax\in\mo$ must hold. 
This can be accomplished for any given $\pr$
by 
choosing an arbitrary fixed $c\in(0,0.5)$ and by setting $\pr_{\mathit{adj}}(\tax) := c * \pr(\tax)$ for all $\tax \in \mo$. Observe that this adjustment does not affect the relative probabilities in that $\pr_{\mathit{adj}}(\tax)/\pr_{\mathit{adj}}(\tax') = k$ whenever $\pr(\tax)/\pr(\tax') = k$, i.e., no information is lost in the sense that the fault probability order of components will remain invariant.

To effect that diagnoses of minimum cardinality (instead of maximal probability) are preferred by RBF-HS, the probability model must satisfy $\pr(\tax) := c$ for all $\tax\in\mo$ for some arbitrary fixed 
$c \in (0,0.5)$. Note, this is equivalent to defining $\pr(\node) := 1/|\node|$ for all nodes $\node$. 

\noindent\textbf{Trivial Cases.}
At the beginning (line~\ref{algoline:rbfhs:initialize_mD_mC_f}), RBF-HS initializes the solution list of found minimal diagnoses $\mD$ and the list of already computed minimal conflicts $\mC$.
Then, two trivial cases are checked, i.e., if no diagnoses exist 
(lines~\ref{algoline:rbfhs:mc=emptyset}--\ref{algoline:rbfhs:return_mD_1}) or 
the empty set is the only diagnosis (lines~\ref{algoline:rbfhs:mc='no_conflict'}--\ref{algoline:rbfhs:return_mD_2}) for $\dpi$. Note, the former case applies iff $\emptyset$ is a conflict for $\dpi$, which implies that $\mo\setminus\emptyset = \mo$ is not a diagnosis for $\dpi$ by the Duality Property (cf.\ Sec.~\ref{sec:basics}), which in turn means that no diagnosis can exist since diagnoses are subsets of $\mo$ and each superset of a diagnosis must be a diagnosis as well (weak fault model, cf.\ Sec.~\ref{sec:basics}). The latter case holds iff there is no conflict at all for $\dpi$, i.e., in particular, $\mo$ is not a conflict, which is why $\mo\setminus\mo = \emptyset$ is a diagnosis by the Duality Property, and consequently no other \emph{minimal} diagnosis can exist. 

If none of these trivial cases is given, the call of \textsc{findMinConflict} (line~\ref{algoline:rbfhs:findMinConflict}) returns a non-empty minimal conflict $\mc$ (line~\ref{algoline:rbfhs:add_mc_to_mC} is reached), which entails by the Hitting Set Property (cf.\ Sec.~\ref{sec:basics}) that a non-empty (minimal) diagnosis will exist. For later reuse (note: conflict computation is an expensive operation), $\mc$ is added to the computed conflicts $\mC$, and then the recursive sub-procedure RBF-HS' is called (line~\ref{algoline:rbfhs:call_RBFHS'}). The arguments passed to RBF-HS' are the root node $\emptyset$, its $f$-value, and the initial bound set to $-\infty$.

\noindent\textbf{Recursion: Abstract View.} 
For better understanding, it is instructive to look upon RBF-HS' as a succession of the following blocks:
\begin{itemize}[noitemsep]
	\item node labeling (line~\ref{algoline:rbfhs':label}),
	\item node elimination or addition to solutions (lines~\ref{algoline:rbfhs':if_nonmin}--\ref{algoline:rbfhs':return_after_valid}), 
	\item node expansion (line~\ref{algoline:rbfhs':expand}),
	\item node cost inheritance (lines~\ref{algoline:rbfhs':for_node_in_childnodes}--\ref{algoline:rbfhs':F(n_i)_gets_f(n_i)}),
	\item child node preparation (lines~\ref{algoline:rbfhs':if_|childnodes|=1}--\ref{algoline:rbfhs':sortDecreasingByF}), and
	\item recursive child node exploration (lines~\ref{algoline:rbfhs':getBestChild_1}--\ref{algoline:rbfhs':return_F(n)}).
\end{itemize}

\noindent\textbf{Recursion: Principle.} 
The basic principle of the recursion (RBF-HS') is 
to always explore the open node (initially, only the root node $\emptyset$) with highest $F$-value (initially, $F$-values are $f$-values) in a depth-first manner, until the best node in the currently explored subtree has 
a lower $F$-value
than the globally best alternative node (whose $F$-value is always stored by $\bnd$). Then backtrack and propagate the best $F$-value among all child nodes up at each backtracking step. Based on their latest known $F$-value, the child nodes at each tree level are re-sorted in best-first order of $F$-value. When re-exploring an already explored, but later forgotten, subtree, the $F$-value of nodes in this subtree is, if necessary, 
updated through an inheritance from parent to children (cf.\ \cite{korf1992linear}). In this vein, 
a re-learning of already learned backed-up $F$-values, and thus repeated and redundant work, is avoided. Exploring a node $\node$ in RBF-HS means labeling $\node$ and assigning it to 
the set of computed minimal diagnoses (collection $\mD$) if the label is $\valid$, and to discard $\node$ (no assignment to any collection) in case it is labeled $\closed$. In both these cases, the backed up $F$-value of $\node$ is set to $-\infty$, which prevents the algorithm to be misled in prospective iterations by good $F$-values of these already explored nodes.
If $\node$'s label is a minimal conflict $\mc$, then $|\mc|$ child nodes $\{\node\cup \{c\}|c\in\mc\}$ are generated and recursively explored.
This recursive backtracking search is executed until either $\mD$ comprises the desired number $\ld$ of minimal diagnoses or the hitting set tree has been explored in its entirety.

\noindent\textbf{Sub-Procedures.}
The workings of the sub-procedures called throughout RBF-HS are:
\begin{itemize}[noitemsep]
	\item $\textsc{findMinConflict}(\dpi)$ receives a DPI $\dpi=\tuple{\mo,\mb,\Tp,\Tn}$ and outputs a minimal conflict $\mc \subseteq \mo$ if one exists, and 'no conflict' else. A well-known algorithm that can be used to implement this function is \textsc{QuickXplain} \cite{junker04,rodler2020qx}.
	\item $\textsc{add}(x,L)$ takes an object $x$ and a list of objects $L$ as inputs, and returns the list obtained by appending the element $x$ to the end of the list $L$.
	\item $\textsc{addDummyNode}(L)$ takes a list of nodes $L$, appends an artificial node $\node$ with $f(\node) := -\infty$ to $L$, and returns the result.
	\item $\textsc{getandDeleteFirstNode}(L)$ accepts a sorted list $L$, deletes the first element from $L$ and returns this deleted element.
	\item $\textsc{getFirstNode}(L)$ accepts a sorted list $L$ and returns $L$'s first element.
	\item $\textsc{sortDecreasingByF}(L)$ accepts a list of nodes $L$, sorts $L$ in decending order of $F$-value, and returns the resulting sorted list.
	\item $\textsc{insertSortedByF}(\node,L)$ accepts a node $\node$ and a list of nodes $L$ sorted by $F$-value, and inserts $\node$ into $L$ in a way the sorting of $L$ by $F$-value is preserved.
\end{itemize}

\noindent Finally, the \textsc{label} function can be seen as a series of the following blocks:
\begin{itemize}[noitemsep]
	\item \emph{non-minimality} check (lines~\ref{algoline:label:non-min_crit_start}--\ref{algoline:label:non-min_crit_end}),
	\item \emph{reuse label} check (lines~\ref{algoline:label:reuse_start}--\ref{algoline:label:reuse_end}), and
	\item \emph{compute label} operations (lines~\ref{algoline:label:findMinConflict}--\ref{algoline:label:return_new_cs}).
\end{itemize}
Note that this \textsc{label} function of RBF-HS' is equal to the one used in Reiter's HS-Tree \cite{Reiter87},
except that the \emph{duplicate} check is obsolete in RBF-HS'. The reason for this is that there cannot ever be any duplicate (i.e., set-equal) nodes in memory at the same time during the execution of RBF-HS. This holds because, for all potential duplicates $\node_i,\node_j$, we must have $|\node_i|=|\node_j|$, but equal-sized nodes must be siblings (depth-first tree exploration) which is why $\node_i$ and $\node_j$ must contain $|\node_i|-1$ equal elements
(same path up to the parent of $\node_i,\node_j$) and one necessarily different element (label of edge pointing from parent to $\node_i$ and $\node_j$, respectively).

\noindent\textbf{Properties.} RBF-HS is sound, complete and best-first, and allows to compute an arbitrary fixed number of diagnoses while requiring only linear memory:\footnote{For more details and a proof of this theorem, see the extended version of this paper at http://isbi.aau.at/ontodebug/publications.} 
\begin{thm}\label{thm:correctness}
	Let $\dpi = \langle\mo,\mb,\Tp,\Tn\rangle$ be a DPI and let \textsc{findMinConflict} be a sound and complete method for conflict computation, i.e., given $\dpi$, it outputs a minimal conflict for $\dpi$ if a minimal conflict exists, and 'no conflict' otherwise.
	RBF-HS is sound, complete and best-first, i.e., it 
	computes \emph{all} and \emph{only} minimal diagnoses for $\dpi$ \emph{in descending order} \emph{of probability} as per the cost-adjusted probability measure $\pr$. Further, given a fixed number $\ld$ of diagnoses to be computed, RBF-HS requires space in $O(|\mo|)$.
\end{thm}

\begin{figure}[th!]
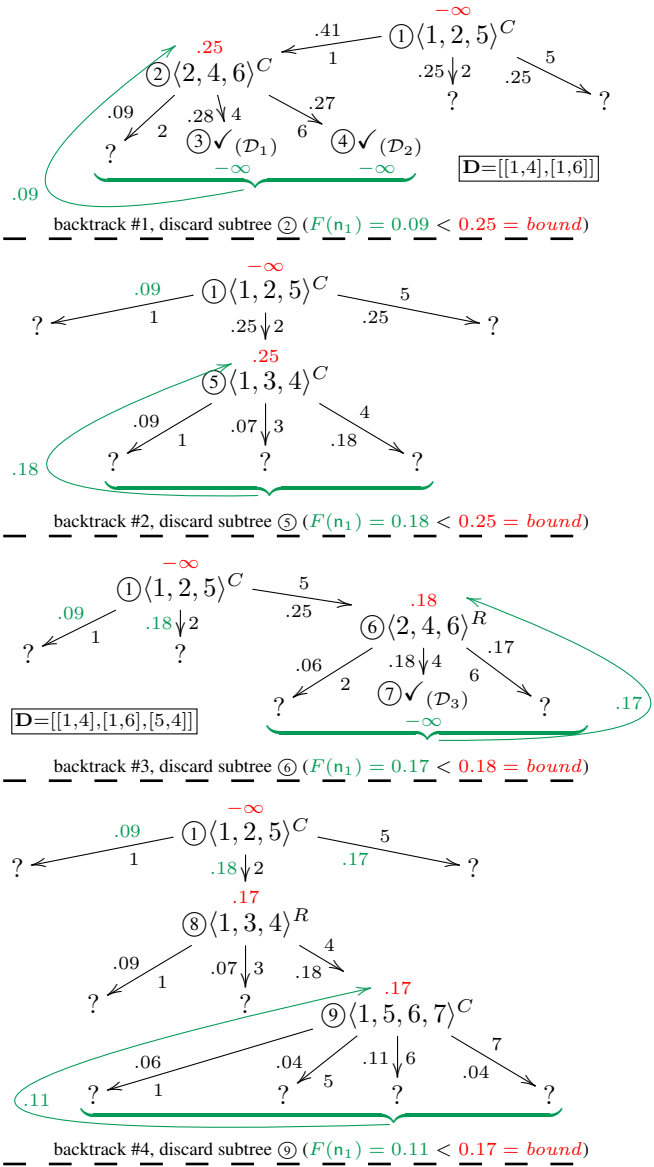

	\centering
	\begin{minipage}[c]{\columnwidth}
		\xygraph{
			!{<0cm,0cm>;<1cm,0cm>:<0cm,-1cm>::}
			!{(-1,0) }*+{ \stackrel{\textcolor{red}{-\infty}}{\textcircled{\scriptsize 1}\langle1,2,5\rangle^{C}} }="n1"
			!{(-4.2,.5) }*+{ \stackrel{\textcolor{red}{.25}}{\textcircled{\scriptsize 2}\langle2,4,6\rangle^{C}} }="n21"
			"n1":"n21"_{.41}^{1}
			!{(-1,1) }*+{? }="n22"
			"n1":"n22"^(0.6){2}_(0.6){.25}
			!{(1,1) }*+{ ? }="n23"
			"n1":"n23"_{.25}^(0.6){5}
			!{(-4.5,.16) }*+{ \phantom{.} }="n2a"
			!{(-5.5,1.7) }*+{ ? }="n31"
			"n21":"n31"_(0.7){.09}^(0.6){2}
			!{(-3.9,1.7) }*+{ \underset{\color{ForestGreen}-\infty}{\textcircled{\scriptsize 3}\checkmark_{(\md_1)}} }="n32"
			"n21":"n32"_(0.55){.28}^(0.6){4}
			!{(-2,1.7) }*+{ \underset{\color{ForestGreen}-\infty}{\textcircled{\scriptsize 4}\checkmark_{(\md_2)}} }="n33"
			"n21":"n33"^(0.6){.27}_(0.6){6}
			!{(-3.6,2.05) }*+{\color{ForestGreen}\underbrace{\phantom{\qquad\qquad\qquad\qquad\qquad\qquad}}  }="n41"
			!{(-0,1.9) }*+{ \boxed{\scriptstyle \mD = [ [1,4],[1,6] ] }}="key"
			!{(-3.6,2.15) }*+{\phantom{.}}="n51"
			"n51" :@/^2.5cm/^{\textcolor{ForestGreen}{.09}}@[ForestGreen] "n2a"
		}
		\vspace{2pt}
		\begin{center}
			\scriptsize 
			backtrack \#1, discard subtree $\textcircled{\scalebox{.7}{2}}$ (${\color{ForestGreen}F(\node_1) = 0.09} < {\color{red}0.25 = \bnd}$)
		\end{center}
		\vspace{-10pt}
		
		\hdashrule[0.5ex]{\columnwidth}{1pt}{3mm} 
	\end{minipage}
	
	\begin{minipage}[c]{\columnwidth}
		\xygraph{
			!{<0cm,0cm>;<1cm,0cm>:<0cm,-1.2cm>::}
			!{(-2,0) }*+{ \stackrel{\textcolor{red}{-\infty}}{\textcircled{\scriptsize 1}\langle1,2,5\rangle^{C}} }="n1"
			!{(-5,.5) }*+{ ? }="n21"
			"n1":"n21"_{\textcolor{ForestGreen}{.09}}^{1}
			!{(-2,1) }*+{ \stackrel{\textcolor{red}{.25}}{\textcircled{\scriptsize 5}\langle1,3,4\rangle^{C}} }="n22"
			"n1":"n22"^(0.5){2}_(0.5){.25}
			!{(1,.5) }*+{ ? }="n23"
			"n1":"n23"_{.25}^(0.6){5}
			!{(-2.3,0.87) }*+{ \phantom{.} }="n2a"
			!{(-4,2) }*+{ ? }="n31"
			"n22":"n31"_(0.7){.09}^(0.6){1}
			!{(-2,2) }*+{ ? }="n32"
			"n22":"n32"_(0.6){.07}^(0.6){3}
			!{(0,2) }*+{ ? }="n33"
			"n22":"n33"^(0.6){4}_(0.6){.18}
			!{(-1.95,2.29) }*+{\color{ForestGreen}\underbrace{\phantom{\;\quad\quad\qquad\qquad\qquad\qquad\qquad}}  }="n41"
			!{(-1.95,2.37) }*+{\phantom{.}}="n51"
			"n51" :@/^2.8cm/^{\textcolor{ForestGreen}{.18}}@[ForestGreen] "n2a"
		}
		\vspace{2pt}
		\begin{center}
			\scriptsize 
			backtrack \#2, discard subtree $\textcircled{\scalebox{.7}{5}}$ (${\color{ForestGreen}F(\node_1) = 0.18} < {\color{red}0.25 = \bnd}$)
		\end{center}
		\vspace{-10pt}
		
		\hdashrule[0.5ex]{\columnwidth}{1pt}{3mm} 
	\end{minipage}
	
	\begin{minipage}[c]{\columnwidth}
		\xygraph{
			!{<0cm,0cm>;<1cm,0cm>:<0cm,-1cm>::}
			!{(-2,0) }*+{ \stackrel{\textcolor{red}{-\infty}}{\textcircled{\scriptsize 1}\langle1,2,5\rangle^{C}} }="n1"
			!{(1.2,.5) }*+{ \stackrel{\textcolor{red}{.18}}{\textcircled{\scriptsize 6}\langle2,4,6\rangle^{R}} }="n21"
			"n1":"n21"_{.25}^{5}
			!{(-2,1) }*+{? }="n22"
			"n1":"n22"^(0.6){2}_(0.6){\textcolor{ForestGreen}{.18}}
			!{(-4,1) }*+{ ? }="n23"
			"n1":"n23"_(0.65){\textcolor{ForestGreen}{.09}}^(0.6){1}
			!{(1.62,.2) }*+{ \phantom{.} }="n2a"
			!{(-0.7,1.7) }*+{ ? }="n31"
			"n21":"n31"_(0.7){.06}^(0.6){2}
			!{(1.2,1.7) }*+{ \underset{\color{ForestGreen}-\infty}{\textcircled{\scriptsize 7}\checkmark_{(\md_3)}} }="n32"
			"n21":"n32"_(0.5){.18}^(0.5){4}
			!{(2.8,1.7) }*+{ ? }="n33"
			"n21":"n33"^(0.5){.17}_(0.5){6}
			!{(-3,1.9) }*+{ \boxed{\scriptstyle \mD = [ [1,4],[1,6],[5,4] ] }}="key"
			!{(1.25,2.05) }*+{\color{ForestGreen}\underbrace{\phantom{\qquad\qquad\qquad\qquad\qquad\qquad}}  }="n41"
			!{(1.25,2.15) }*+{\phantom{.}}="n51"
			"n51" :@/_2.8cm/^{\textcolor{ForestGreen}{.17}}@[ForestGreen] "n2a"
		}
		\vspace{2pt}
		\begin{center}
			\scriptsize 
			backtrack \#3, discard subtree $\textcircled{\scalebox{.7}{6}}$ (${\color{ForestGreen}F(\node_1) = 0.17} < {\color{red}0.18 = \bnd}$)
		\end{center}
		\vspace{-10pt}
		
		\hdashrule[0.5ex]{\columnwidth}{1pt}{3mm} 
	\end{minipage}
	
	\begin{minipage}[c]{\columnwidth}
		\xygraph{
			!{<0cm,0cm>;<1cm,0cm>:<0cm,-1.2cm>::}
			!{(-2,0) }*+{ \stackrel{\textcolor{red}{-\infty}}{\textcircled{\scriptsize 1}\langle1,2,5\rangle^{C}} }="n1"
			!{(-5,.5) }*+{ ? }="n21"
			"n1":"n21"_{\textcolor{ForestGreen}{.09}}^{1}
			!{(-2,1) }*+{ \stackrel{\textcolor{red}{.17}}{\textcircled{\scriptsize 8}\langle1,3,4\rangle^{R}} }="n22"
			"n1":"n22"^(0.5){2}_(0.5){\textcolor{ForestGreen}{.18}}
			!{(1,.5) }*+{ ? }="n23"
			"n1":"n23"_{\textcolor{ForestGreen}{.17}}^(0.6){5}
			!{(-0.2,1.8) }*+{ \phantom{.} }="n2a"
			!{(-4,2) }*+{ ? }="n31"
			"n22":"n31"_(0.7){.09}^(0.6){1}
			!{(-2,2) }*+{ ? }="n32"
			"n22":"n32"_(0.6){.07}^(0.6){3}
			!{(0,2) }*+{ \stackrel{\textcolor{red}{.17}}{\textcircled{\scriptsize 9}\langle1,5,6,7\rangle^{C}} }="n33"
			"n22":"n33"^(0.5){4}_(0.5){.18}
			!{(-1,3.22) }*+{\color{ForestGreen}\aunderbrace[l8D4r]{\phantom{\hspace{180pt}}}  }="n4a"
			!{(-4,3) }*+{ ? }="n41"
			"n33":"n41"^(0.8){1}_(0.8){.06}
			!{(-1.5,3) }*+{ ? }="n42"
			"n33":"n42"^(0.7){5}_(0.8){.04}
			!{(0,3) }*+{ ? }="n43"
			"n33":"n43"^(0.6){6}_(0.6){.11}
			!{(2,3) }*+{ ? }="n44"
			"n33":"n44"^(0.6){7}_(0.6){.04}
			!{(0.05,3.33) }*+{\phantom{.}}="n5a"
			"n5a" :@/^5cm/_{\textcolor{ForestGreen}{.11}}@[ForestGreen] "n2a"
		}
		\vspace{2pt}
		\begin{center}
			\scriptsize 
			backtrack \#4, discard subtree $\textcircled{\scalebox{.7}{9}}$ (${\color{ForestGreen}F(\node_1) = 0.11} < {\color{red}0.17 = \bnd}$)
		\end{center}
		\vspace{-10pt}
		
		\hdashrule[0.5ex]{\columnwidth}{1pt}{3mm} 
	\end{minipage}
	
	\caption{RBF-HS executed on example DPI (part I).}
	\label{fig:rbfhs_example_part1}
\end{figure}

\begin{figure}[th!]
	\centering
	
	\begin{minipage}[c]{\columnwidth}
		\xygraph{
			!{<0cm,0cm>;<1cm,0cm>:<0cm,-1.2cm>::}
			!{(-2,0) }*+{ \stackrel{\textcolor{red}{-\infty}}{\textcircled{\scriptsize 1}\langle1,2,5\rangle^{C}} }="n1"
			!{(-5,.5) }*+{ ? }="n21"
			"n1":"n21"_{\textcolor{ForestGreen}{.09}}^{1}
			!{(-2,1) }*+{ \stackrel{\textcolor{red}{.17}}{\textcircled{\scriptsize 8}\langle1,3,4\rangle^{R}} }="n22"
			"n1":"n22"^(0.5){2}_(0.5){\textcolor{ForestGreen}{.18}}
			!{(1,.5) }*+{ ? }="n23"
			"n1":"n23"_{\textcolor{ForestGreen}{.17}}^(0.6){5}
			!{(-2.3,0.87) }*+{ \phantom{.} }="n2a"
			!{(-4,2) }*+{ ? }="n31"
			"n22":"n31"_(0.7){.09}^(0.6){1}
			!{(-2,2) }*+{ ? }="n32"
			"n22":"n32"_(0.6){.07}^(0.6){3}
			!{(0,2) }*+{ ? }="n33"
			"n22":"n33"^(0.6){4}_(0.6){\textcolor{ForestGreen}{.11}}
			!{(-1.95,2.29) }*+{\color{ForestGreen}\underbrace{\phantom{\;\quad\quad\qquad\qquad\qquad\qquad\qquad}}  }="n41"
			!{(-1.95,2.37) }*+{\phantom{.}}="n51"
			"n51" :@/^2.8cm/^{\textcolor{ForestGreen}{.11}}@[ForestGreen] "n2a"
		}
		\vspace{2pt}
		\begin{center}
			\scriptsize 
			backtrack \#5, discard subtree $\textcircled{\scalebox{.7}{8}}$ (${\color{ForestGreen}F(\node_1) = 0.11} < {\color{red}0.17 = \bnd}$)
		\end{center}
		\vspace{-10pt}
		
		\hdashrule[0.5ex]{\columnwidth}{1pt}{3mm} 
	\end{minipage}
	
	\begin{minipage}[c]{\columnwidth}
		\xygraph{
			!{<0cm,0cm>;<1cm,0cm>:<0cm,-1cm>::}
			!{(-2,0) }*+{ \stackrel{\textcolor{red}{-\infty}}{\textcircled{\scriptsize 1}\langle1,2,5\rangle^{C}} }="n1"
			!{(1.2,.5) }*+{ \stackrel{\textcolor{red}{.11}}{\textcircled{\tiny 10}\langle2,4,6\rangle^{R}} }="n21"
			"n1":"n21"_{\textcolor{ForestGreen}{.17}}^{5}
			!{(-2,1) }*+{? }="n22"
			"n1":"n22"^(0.6){2}_(0.6){\textcolor{ForestGreen}{.11}}
			!{(-4,1) }*+{ ? }="n23"
			"n1":"n23"_(0.65){\textcolor{ForestGreen}{.09}}^(0.6){1}
			!{(2.9,1.4) }*+{ \phantom{.} }="n2a"
			!{(-0.7,1.7) }*+{ ? }="n31"
			"n21":"n31"_(0.7){.06}^(0.6){2}
			!{(1.2,1.7) }*+{ \underset{\color{ForestGreen}-\infty}{\textcircled{\tiny 11}\times_{(=\md_3)}^*} }="n32"
			"n21":"n32"_(0.5){.18}^(0.5){4}
			!{(2.8,1.7) }*+{ \stackrel{\textcolor{red}{.11}}{\textcircled{\tiny 12}\langle1,3,4\rangle^{R}} }="n33"
			"n21":"n33"^(0.5){.17}_(0.5){6}
			!{(-0.2,3) }*+{ ? }="n41"
			"n33":"n41"^(0.7){1}_(0.8){.06}	
			!{(1.6,3) }*+{ ? }="n42"
			"n33":"n42"^(0.7){3}_(0.8){.04}
			!{(3.2,3) }*+{ \underset{\color{ForestGreen}-\infty}{\textcircled{\tiny 13}\times_{(\supset\md_3)}} }="n43"
			"n33":"n43"^(0.5){.11}_(0.5){4}						
			!{(1.8,3.35) }*+{\color{ForestGreen}\aunderbrace[l7D4r]{\phantom{\;\qquad\qquad\qquad\qquad\qquad\qquad}}  }="n41"
			!{(2.2,3.5) }*+{\phantom{.}}="n51"
			"n51" :@/_2.1cm/_(0.82){\textcolor{ForestGreen}{.06}}@[ForestGreen] "n2a"
		}
		\vspace{2pt}
		\begin{center}
			\scriptsize 
			backtrack \#6, discard subtree $\textcircled{\scalebox{.57}{12}}$ (${\color{ForestGreen}F(\node_1) = 0.06} < {\color{red}0.11 = \bnd}$)
		\end{center}
		\vspace{-10pt}
		
		\hdashrule[0.5ex]{\columnwidth}{1pt}{3mm} 
	\end{minipage}
	
	\begin{minipage}[c]{\columnwidth}
		\xygraph{
			!{<0cm,0cm>;<1cm,0cm>:<0cm,-1cm>::}
			!{(-2,0) }*+{ \stackrel{\textcolor{red}{-\infty}}{\textcircled{\scriptsize 1}\langle1,2,5\rangle^{C}} }="n1"
			!{(1.2,.5) }*+{ \stackrel{\textcolor{red}{.11}}{\textcircled{\tiny 10}\langle2,4,6\rangle^{R}} }="n21"
			"n1":"n21"_{\textcolor{ForestGreen}{.17}}^{5}
			!{(-2,1) }*+{? }="n22"
			"n1":"n22"^(0.6){2}_(0.6){\textcolor{ForestGreen}{.11}}
			!{(-4,1) }*+{ ? }="n23"
			"n1":"n23"_(0.65){\textcolor{ForestGreen}{.09}}^(0.6){1}
			!{(1.75,0.15) }*+{ \phantom{.} }="n2a"
			!{(-0.7,1.7) }*+{ ? }="n31"
			"n21":"n31"_(0.7){.06}^(0.6){2}
			!{(1.2,1.7) }*+{ \underset{\color{ForestGreen}-\infty}{\textcircled{\tiny 11}\times_{(=\md_3)}} }="n32"
			"n21":"n32"_(0.5){.18}^(0.5){4}
			!{(2.8,1.7) }*+{ ? }="n33"
			"n21":"n33"^(0.65){6}_(0.5){\textcolor{ForestGreen}{.06}}					
			!{(1.05,2.05) }*+{\color{ForestGreen}\aunderbrace[l4D5r]{\phantom{\quad\qquad\qquad\qquad\qquad\qquad}}  }="n41"
			!{(0.8,2.2) }*+{\phantom{.}}="n51"
			"n51" :@/_2cm/_(0.65){\textcolor{ForestGreen}{.06}}@[ForestGreen] "n2a"
		}
		\vspace{2pt}
		\begin{center}
			\scriptsize 
			backtrack \#7, discard subtree $\textcircled{\scalebox{.57}{10}}$ (${\color{ForestGreen}F(\node_1) = 0.06} < {\color{red}0.11 = \bnd}$)
		\end{center}
		\vspace{-10pt}
		
		\hdashrule[0.5ex]{\columnwidth}{1pt}{3mm} 
	\end{minipage}
	
	\begin{minipage}[c]{\columnwidth}
		\xygraph{
			!{<0cm,0cm>;<1cm,0cm>:<0cm,-1.2cm>::}
			!{(-2,0) }*+{ \stackrel{\textcolor{red}{-\infty}}{\textcircled{\scriptsize 1}\langle1,2,5\rangle^{C}} }="n1"
			!{(-5,.5) }*+{ ? }="n21"
			"n1":"n21"_{\textcolor{ForestGreen}{.09}}^{1}
			!{(-2,1) }*+{ \stackrel{\textcolor{red}{.09}}{\textcircled{\tiny 14}\langle1,3,4\rangle^{R}} }="n22"
			"n1":"n22"^(0.5){\textcolor{ForestGreen}{.11}}_(0.5){2}
			!{(1,.5) }*+{ ? }="n23"
			"n1":"n23"_{\textcolor{ForestGreen}{.06}}^(0.6){5}
			!{(-0.2,1.8) }*+{ \phantom{.} }="n2a"
			!{(-1.65,0.53) }*+{ \phantom{.} }="n1.5a"
			!{(-0.8,1.3) }*+{ \phantom{.} }="n2.5a"
			!{(-4,2) }*+{ ? }="n31"
			"n22":"n31"_(0.7){.09}^(0.6){1}
			!{(-2,2) }*+{ ? }="n32"
			"n22":"n32"_(0.6){.07}^(0.6){3}
			!{(0,2) }*+{ \stackrel{\textcolor{red}{.09}}{\textcircled{\tiny 15}\langle1,5,6,7\rangle^{C}} }="n33"
			"n22":"n33"^(0.5){\textcolor{brown}{.11}}_(0.5){4}
			%
			!{(-4,3) }*+{ ? }="n41"
			"n33":"n41"^(0.8){1}_(0.8){.06}
			!{(-1.5,3) }*+{ ? }="n42"
			"n33":"n42"^(0.7){5}_(0.8){.04}
			!{(0,3) }*+{ \textcircled{\tiny 16}\checkmark_{(\md_4)} }="n43"
			"n33":"n43"^(0.6){6}_(0.6){.11}
			!{(2,3) }*+{ ? }="n44"
			"n33":"n44"^(0.6){7}_(0.6){.04}
			!{(1.6,1.3) }*+{ \boxed{\scriptstyle \mD = [ [1,4],[1,6],[5,4],[2,4,6] ] }}="key"
			"n1.5a" :@/^0.4cm/^(0.7){\text{\textcolor{brown}{inherit}}}@[brown] "n2.5a"
		}
		\vspace{2pt}
		\begin{center}
			\scriptsize 
			exit procedure ($|\mD| = 4 \geq 4 = \ld$) $\quad\Rightarrow\quad$ return $\mD$
		\end{center}
		\vspace{-10pt}
		
		\hdashrule[0.5ex]{\columnwidth}{1pt}{3mm} 
	\end{minipage}

	\caption{RBF-HS executed on example DPI (part II).}
	\label{fig:rbfhs_example_part2}
\end{figure}

\noindent\textbf{Illustration.}
We next visualize the workings of RBF-HS:
\begin{example} \hspace{-1em}\emph{(RBF-HS)}\quad\label{ex:RBF-HS}
	\noindent\emph{Inputs.} Consider a defective system 
	described by 
	$\dpi := \langle\mo,\mb,\Tp,\Tn\rangle$, where $\mo = \{\tax_1,\dots,\tax_7\}$ 
	and no background knowledge or 
	positive and negative measurements are given,
	i.e., $\mb,\Tp,\Tn = \emptyset$.
	Let $\langle \pr(\tax_1), \dots, \pr(\tax_7)\rangle := \langle .26,.18,.21$, $.41,.18,.40,.18\rangle$ (note: $\pr$ is already cost-adjusted, cf.\ Sec.~\ref{sec:algo}). Further, let all minimal conflicts for $\dpi$ be $\langle\tax_1,\tax_2,\tax_5\rangle$, $\langle\tax_2,\tax_4,\tax_6\rangle$, $\langle\tax_1,\tax_3,\tax_4\rangle$, and $\langle\tax_1,\tax_5,\tax_6, \tax_7\rangle$. Assume we want to use RBF-HS to find the $\ld := 4$ most probable diagnoses for $\dpi$. 
	To this end, $\dpi$, $\pr$ and $\ld$ are passed to RBF-HS (Alg.~\ref{algo:RBF_HS}) as input arguments.\vspace{2pt}
	
	\noindent\emph{Illustration (Figures).} The way of proceeding of RBF-HS is depicted by Figures~\ref{fig:rbfhs_example_part1} and \ref{fig:rbfhs_example_part2}, where the following notation is used. 
	Axioms $\tax_i$ are simply referred to by $i$ (in node and edge labels). Numbers $\textcircled{\scriptsize k}$ indicate the chronological node labeling (expansion) order. Recall that nodes in Alg.~\ref{algo:RBF_HS} are sets of (integer) edge labels along tree branches. E.g., node $\textcircled{\scriptsize 9}$ in Fig.~\ref{fig:rbfhs_example_part1} corresponds to the node $\node = \{\tax_2,\tax_4\}$, i.e., to the assumption that components $c_2,c_4$ are at fault whereas all others are working properly. The probability $\pr(\node)$ (i.e., the original $f$-value) of a node $\node$ is shown by the black number from the interval $(0,1)$ that labels the edge pointing to $\node$, e.g., the cost of node $\textcircled{\scriptsize 9}$ is $0.18$. 
	We tag minimal conflicts $\tuple{\dots}$ that label internal nodes by $^C$ if they are freshly computed (\emph{expensive}; \textsc{findMinConflict} call, line~\ref{algoline:label:findMinConflict}),
	and 
	by $^R$ if they result from a reuse of some already computed and stored (see list $\mC$ in Alg.~\ref{algo:RBF_HS}) minimal conflict
	(\emph{cheap}; reuse label check; lines~\ref{algoline:label:reuse_start}--\ref{algoline:label:reuse_end}).
	%
	Leaf nodes are labeled as follows:
	``$?$'' is used for open (i.e., generated, but not yet labeled) nodes;
	$\checkmark_{(\md_i)}$ for a node 
	labeled $\valid$, i.e., a minimal diagnosis named $\md_i$, that is not yet stored in $\mD$;
	$\times_{(\mathit{Expl})}$ for a node labeled $\closed$, i.e., one that constitutes a non-minimal diagnosis or a diagnosis that has already been found and stored in $\mD$; $\mathit{Expl}$ is an explanation for the non-minimality in the former, and for the redundancy of node in the latter case, i.e., $\mathit{Expl}$ names a minimal diagnosis in $\mD$ that is a proper subset of the node, or it names a diagnosis in $\mD$ which is equal to node, respectively. Whenever a new diagnosis is 
	added to $\mD$ (line~\ref{algoline:rbfhs:add_node_to_mD}), this is displayed in the figures by a box that shows the current state of $\mD$. 
	For each expanded node, the value of the $\bnd$ variable relevant to the subtree rooted at this node is denoted by a red-colored value above the node. By green color, we show the backed-up $F$-value returned in the course of each backtracking step 
	(i.e., the best known probability of any node in the respective subtree). Further, $f$-values that have been updated by backed-up $F$-values are signalized by {green-colored} edge labels, see, e.g., in Fig.~\ref{fig:rbfhs_example_part1}, the left edge emanating from the root node of the tree has been reduced from $0.41$ ($f$-value) to $0.09$ ($F$-value) after the first backtrack. Finally, $F$-values of parents inherited by child nodes 
	(line~\ref{algoline:rbfhs':F(n_i)_gets_min}) are indicated by brown color, see the edge between node $\textcircled{\tiny 14}$ and node $\textcircled{\tiny 15}$ in Fig.~\ref{fig:rbfhs_example_part2}.\vspace{2pt}
	
	\noindent\emph{Discussion and Remarks.}
	Initially, RBF-HS starts with an empty root node, labels it with the minimal conflict $\tuple{1,2,5}$ at step $\textcircled{\scriptsize 1}$, generates the three corresponding child nodes $\{1\}, \{2\}, \{5\}$ shown by the edges originating from the root node, and recursively processes the best child node 
	(left edge, $f$-value $0.41$) at step $\textcircled{\scriptsize 2}$. The $\bnd$ for the subtree rooted at node $\textcircled{\scriptsize 2}$ corresponds to the best edge label ($F$-value) of any open node other than node $\textcircled{\scriptsize 2}$, which is $0.25$ in this 
	case. In a similar manner, the next recursive step is taken in that the best child node of node $\textcircled{\scriptsize 2}$ with an $F$-value not less than $\bnd = 0.25$ is processed. This leads to the labeling of node $\{1,4\}$ with $F$-value $0.28 \geq \bnd$ at step $\textcircled{\scriptsize 3}$, which reveals the first (provenly most probable) diagnosis $\md_1 := [1,4]$ with $\pr(\md_1) = 0.28$, which is added to the solution list $\mD$. Note that $-\infty$ is at the same time returned for node $\textcircled{\scriptsize 3}$. After the next node has been processed and the second-most-probable minimal diagnosis $\md_2 := [1,6]$ 
	with $\pr(\md_2) = 0.27$
	has been detected, the by now best remaining child node of node $\textcircled{\scriptsize 2}$ has an $F$-value of $0.09$ (leftmost node). This value, however, is lower than $\bnd$. Due to the best-first property of RBF-HS, this node is not explored right away because $\bnd$ suggests that there are more promising unexplored nodes elsewhere in the tree which have to be checked first. To keep the memory requirements linear, the current subtree rooted at node $\textcircled{\scriptsize 2}$ is discarded before a new one is examined. Hence, the first backtrack is executed. This involves 
	the storage of the best (currently known) $F$-value of any node in the subtree as the backed-up $F$-value of node $\textcircled{\scriptsize 2}$. This newly ``learned'' $F$-value is signalized by the green number ($0.09$) that by now labels the left edge emanating from the root. 
	Analogously, RBF-HS proceeds for the other nodes, 
	whereas the used $\bnd$ value is always the best value among the $\bnd$ value of the parent and all sibling's $F$-values. 
	Please also observe the $F$-value inheritance that takes place when node $\{2,4\}$ is generated for the third time (node $\textcircled{\tiny 15}$, Fig.~\ref{fig:rbfhs_example_part2}). The reason for this is that the original $f$-value of $\{2,4\}$ is $0.18$ (see top of Fig.~\ref{fig:rbfhs_example_part1}), but the meanwhile ``learned'' $F$-value of its parent $\{2\}$ is $0.11$ and thus smaller.
	This means that $\{2,4\}$ must have already been explored and the \emph{de-facto} probability of any 
	diagnosis in the subtree rooted at $\{2,4\}$ must be less than or equal to $0.11$.\vspace{2pt}
	
	\noindent\emph{Output.} RBF-HS immediately terminates as soon as the $\ld$-th (in this case: fourth) minimal diagnosis $\md_4$ is located and added to $\mD$. The list $\mD$ of minimal diagnoses arranged in descending order of probability $\pr$ is returned.\qed
\end{example}

	\setlength{\tabcolsep}{6pt}
\begin{table}
	\renewcommand\arraystretch{1}
	\scriptsize
	\centering
	\caption{\small Dataset used in the experiments (sorted by 2nd column).}
	\label{tab:dataset}
	\begin{minipage}{0.98\linewidth}
		\begin{tabular}{@{}lrlr@{}} 
			\toprule
			KB $\mo$				& $|\mo|$& expressivity \textsuperscript{\textbf{1)}} 		& \#D/min/max \textsuperscript{\textbf{2)}} \\ \midrule
			Koala (K) 
			& 42 		& $\mathcal{ALCON}^{(D)}$& 10/1/3     \\
			University (U) 
			& 50 		& $\mathcal{SOIN}^{(D)}$& 90/3/4      \\
			IT  
			& 
			140 		& $\mathcal{SROIQ}$& 1045/3/7	  \\
			UNI  
			\phantom{\textsuperscript{\textbf{4)}}}		
			& 
			142 		& $\mathcal{SROIQ}$& 1296/5/6	  \\
			Chemical (Ch) 
			& 144 		& $\mathcal{ALCHF}^{(D)}$& 6/1/3     \\
			MiniTambis (M) 
			\phantom{\textsuperscript{\textbf{4)}}}		
			& 173 		& $\mathcal{ALCN}$ 		& 48/3/3	  \\
			
			Transportation (T) 
			& 
			1300 		& $\mathcal{ALCH}^{(D)}$& 1782/6/9	  \\
			Economy (E) 
			& 1781 		& $\mathcal{ALCH}^{(D)}$& 864/4/8     \\
			DBpedia (D) 
			& 
			7228 		& $\mathcal{ALCHF}^{(D)}$& 7/1/1     \\
			Opengalen (O) 
			& 
			9664		& $\mathcal{ALEHIF}^{(D)}$& 110/2/6     \\
			CigaretteSmokeExposure (Cig) 
			& 
			26548 		& $\mathcal{SI}^{(D)}$& 1566*/4/7*	  \\
			Cton (C) 
			& 
			33203		& $\mathcal{SHF}$& 15/1/5     \\
			\bottomrule
		\end{tabular}
	\end{minipage}
	\renewcommand\arraystretch{1}
	\begin{minipage}{0.98\linewidth}
		\setlength{\tabcolsep}{2pt}
		\begin{tabular}{@{}lp{7.5cm}@{}}
			\textbf{1):} & Description Logic expressivity, cf.\ \cite{DLHandbook};the higher the expressivity, 
			the higher is the complexity of consistency checking (conflict computation).
			\\
			\textbf{2):} & \#D/min/max denotes the number/the min.\ size/the max.\ size of minimal diagnoses for the DPI resulting from each input KB $\mo$. If tagged with a $^*$, a value signifies the number/size determined within 1200sec using HS-Tree. 
		\end{tabular}
	\end{minipage}
\end{table}

\section{Evaluation}
\label{sec:eval}

\noindent\textbf{Dataset.}
As a test dataset for our experiments with RBF-HS 
we used twelve diagnosis problems from the knowledge-base debugging domain (Table~\ref{tab:dataset}) where RBF-HS's features soundness, completeness and best-firstness
are important requirements to diagnosis searches \cite{meilicke2011thesis,Rodler2015phd,Kalyanpur2006a}. These problems were already analyzed in studies conducted by other works, e.g., \cite{Shchekotykhin2012,Kalyanpur2006a,rodler2019KBS_userstudy}, and represent particularly challenging cases in terms of the complexity of consistency checking (e.g., a consistency check for $\mathcal{SROIQ}$, cf.\ third column of Table~\ref{tab:dataset}, is 2-NEXPTIME-complete \cite{grau2008owl}). 
As Table~\ref{tab:dataset} shows, the dataset also covers a spectrum of different problem sizes (number of axioms or components; column 2), logical expressivities (column 3),  
as well as diagnostic structures (number and size of minimal diagnoses; column 4).
Note that every model-based diagnosis problem (according to Reiter's original characterization \cite{Reiter87}) can be represented as a knowledge-base debugging problem \cite{rodler17dx_reducing}, which is why considering knowledge-base debugging problems is without loss of generality.

\noindent\textbf{Experiment Settings.} 
We evaluate RBF-HS in relation to Reiter's HS-Tree \cite{Reiter87}, which is a state-of-the-art 
sound, complete and best-first diagnosis search that is as generally applicable as RBF-HS, due to its independence from the used theorem prover and from the logical language used to describe the diagnosed system.
%

In our experiments, we considered a multitude of different \emph{diagnosis scenarios}. 
A diagnosis scenario is defined by the set of inputs given to Alg.~\ref{algo:RBF_HS}, i.e., by a DPI $\dpi$, a number $\ld$ of minimal diagnoses to be computed, as well as a (cost-adjusted) setting of the component fault probabilities $\pr$.
The DPIs for our tests were defined as $\tuple{\mo,\emptyset,\emptyset,\emptyset}$, one for each $\mo$ in Tab.~\ref{tab:dataset}. That is, the task was to find a minimal set of axioms (faulty components) responsible for the inconsistency of $\mo$, without any background knowledge or measurements initially given (cf.\ Example~\ref{ex:RBF-HS}). For the parameter $\ld$ we used the values $\{2,6,10,20\}$. The fault probability $\pr(\tax)$ of each axiom (component) $\tax \in \mo$
 was specified in a way the diagnosis search returns minimum-cardinality diagnoses first (cf.\ Sec.~\ref{sec:algo}). As a logical theorem prover, we adopted Pellet \cite{Sirin2007}.

To simulate as realistic as possible diagnosis circumstances, where the \emph{actual diagnosis} (i.e., the de-facto faulty axioms) 
needs to be isolated from a set of initial minimal diagnoses, 
we ran five sequential diagnosis \cite{dekleer1987,Shchekotykhin2012} sessions for each diagnosis scenario defined above. At this, a different randomly chosen actual diagnosis was set as the target solution in each session. 
Note, running sequential diagnosis sessions instead of just applying a single diagnosis search execution to the DPIs listed in Table~\ref{tab:dataset} has the additional advantage that multiple diagnosis searches, each for a different (updated) DPI, are executed during one sequential session and flow into the experiment results, which gives us a more representative picture of the algorithms' real performance.
%

A sequential diagnosis session can be conceived of having two alternating phases, that are iterated until a single diagnosis remains: diagnosis search, and measurement conduction. More precisely, the former involves the determination of $\ld$ minimal diagnoses $\mD$ for a given DPI, the latter the computation of an optimal system measurement (to rule out as many spurious diagnoses in $\mD$ as possible), as well as the incorporation of the new 
knowledge resulting from the measurement outcome into the DPI.
Measurement computation is accomplished by means of a \emph{measurement selection function} which gets a set of minimal diagnoses $\mD$ as input, and outputs one system measurement such that any measurement outcome eliminates at least one spurious diagnosis in $\mD$. In our experiments, a measurement was defined as a true-false question to an oracle \cite{Shchekotykhin2014,Shchekotykhin2012,Rodler2015phd,rodler2019KBS_userstudy}, e.g., for a biological KB one such query could be $Q:=\mathsf{Bird} \sqsubseteq \mathsf{FlyingAnimal}$ (``does every bird fly?''). Given a positive (negative) answer, $Q$ is moved to the positive (negative) measurements of the DPI. The new DPI is then used in the next iteration of the sequential diagnosis session. That is, a new set of diagnoses $\mD$ is sought for this updated DPI, an optimal measurement is calculated for $\mD$, and so forth. Once there is only a single minimal diagnosis for a current DPI, the session stops and outputs the remaining diagnosis. To determine measurement outcomes (i.e., to answer generated questions), we used the predefined actual diagnosis, i.e., each question was automatically answered in a way the actual diagnosis was not ruled out.
As a measurement selection function we adopted 
the commonly used \emph{entropy (ENT)} heuristic \cite{dekleer1987,Shchekotykhin2012}, which selects a measurement with highest information gain. 


To sum up: We ran five diagnosis sessions, each searching for a randomly specified minimal diagnosis, for each algorithm among RBF-HS
and HS-Tree, 
for each DPI from Tab.~\ref{tab:dataset}, 
and for each number of diagnoses $\ld \in \{2,6,10,20\}$ to be computed (in each iteration of the session, i.e., at each call of a diagnosis search algorithm). 


\begin{figure}[t]
	\centering
	\includegraphics[width=\columnwidth]{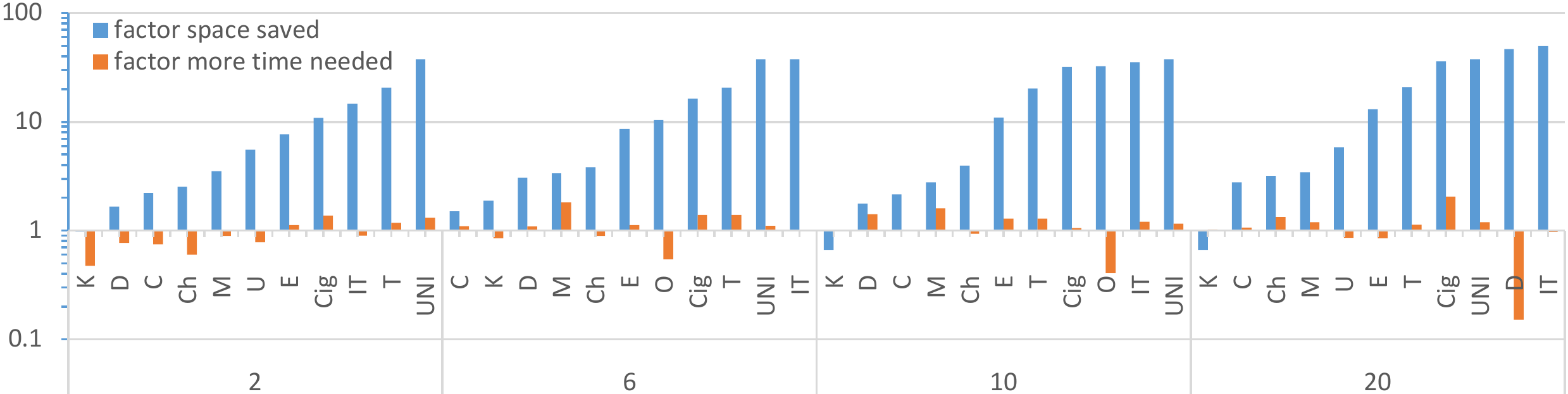}
	\caption{\small Experiment Results: x-axis shows KBs from Table~\ref{tab:dataset} and parameter $\ld \in \{2,6,10,20\}$. y-axis pictures factor of memory saved and factor of more time needed by RBF-HS (vs.\ HS-Tree).}
	\label{fig:results_card_ENT}
\end{figure}

\noindent\textbf{Experiment Results}
are shown by Figure~\ref{fig:results_card_ENT},
which compares the runtime and memory consumption we measured for RBF-HS and HS-Tree averaged over the five performed sessions (note the logarithmic scale).
More specifically, the figure depicts the factor of less memory consumed by RBF-HS (blue bars), as well as the factor of more time needed by RBF-HS (orange bars), in relation to HS-Tree.
That is, blue bars tending upwards (downwards) mean a better (worse) memory behavior of RBF-HS, whereas upwards (downwards) orange bars signify worse (better) runtime of RBF-HS. For instance, a blue bar of height 10 means that HS-Tree required 10 times as much memory as RBF-HS did in the same experiment; or a downwards orange bar representing the value 0.5 indicates that RBF-HS finished the diagnosis search task in half of HS-Tree's runtime. 
Regarding the absolute runtime and memory expenditure (not displayed in Figure~\ref{fig:results_card_ENT}) in the experiments, we measured a min/avg/max runtime of 0.04/24/744sec 
as well as a min/avg/max space consumption of 9/17.5K/1.3M tree nodes.

We make the following observations based on Figure~\ref{fig:results_card_ENT}:\footnote{Note, both RBF-HS and HS-Tree always return exactly the same diagnoses because they provenly have the same features (soundness, completeness, best-firstness). Hence, any savings observed do \emph{not} arise at the cost of losing any theoretical guarantees.}
\begin{enumerate}[itemsep=2pt,label=\emph{(\arabic*)},align=left,leftmargin=0pt,labelwidth=-0.5\parindent]  
\item \emph{More space gained than extra time expended:}  
Whenever the diagnosis problem was non-trivial to solve, RBF-HS trades space favorably for time, i.e., the factor of space saved is higher than the factor of time overhead (blue bar is higher than orange one). 
%
%
\item \emph{Substantial space savings:} Space savings of RBF-HS range from significant to tremendous, and often reach values larger than 10 and up to 50. 
In other words, the memory overhead of HS-Tree compared to RBF-HS for the same diagnostic task reached up to 4900\,\%. 
\item \emph{Often also favorable runtime:} In 40\,\% 
of the cases RBF-HS exhibited even a lower or equal runtime compared with HS-Tree. In fact, the runtime savings achieved by RBF-HS reach values of up to more than 88\,\% (case D,20) while at the same time often saving more than 90\,\% of space.  
Note, also studies comparing classic (non-hitting-set) best-first searches have observed that linear-space approaches can outperform exponential-space ones in terms of runtime. One reason for this is that, at the processing of each node, the management (node insertion and removal) of an exponential-sized priority queue of open nodes requires time linear in the current tree depth \cite{zhang1995performance}. Hence, when the queue management time of HS-Tree outweighs the time for redundant node regenerations expended by RBF-HS, then the latter will outperform the former. 
\item \emph{Performance independent of number of computed diagnoses:} The relative performance of RBF-HS versus HS-Tree appears to be largely independent of the number $\ld$ of computed minimal diagnoses. 
\item \emph{Performance dependent on diagnosis problem:} 
The gain of using RBF-HS instead of HS-Tree gets the larger, the harder the considered diagnosis problem is. This tendency can be clearly seen in Figure~\ref{fig:results_card_ENT} 
where the diagnosis problems on the x-axis are sorted in ascending order of RBF-HS's memory reduction achieved, for each value of $\ld$. Note that roughly the same group of (more difficult / easy to solve) diagnosis problems ranks high / low for all values of $\ld$.

\end{enumerate}  

\section{Conclusions and Future Work}
\label{sec:conclusion}
%
We introduced 
RBF-HS, a \emph{general} (reasoner-independent and logics-independent) diagnosis (or: hitting set) search that computes minimal diagnoses (hitting sets) in a \emph{sound} and \emph{complete} way, and enumerates them in \emph{best-first} order as prescribed by some preference function (e.g., minimum cardinality, maximal probability). In contrast to existing systems in model-based diagnosis, RBF-HS guarantees these three properties under \emph{linear-space memory bounds}.


In experiments on a corpus of real-world diagnosis problems of various size, reasoning complexity, and diagnostic structure, we put RBF-HS to the test on minimum-cardinality diagnosis computation tasks. At this, we compared RBF-HS against HS-Tree, a state-of-the-art sound, complete and best-first hitting set algorithm which is equally general (i.e., reasoner-independent and logics-independent) as RBF-HS.
The results testify that:
\emph{(1)}~RBF-HS achieves significantly higher space savings than time losses in all non-trivial cases, and the performance gains tend to increase with increasing problem size and complexity; 
\emph{(2)}~in many cases, RBF-HS's improvements of memory costs are enormous, reaching savings of up to 98\,\%;
\emph{(3)}~the memory advantages reached by RBF-HS mostly do \emph{not} come at the cost of notable runtime increases;
\emph{(4)}~in four out of ten cases, the runtime of RBF-HS was even lower than that of HS-Tree, and runtime savings reached values of up to more than 88\,\%.

Future work topics 
include \emph{(1)}~further evaluations of RBF-HS, e.g., when used to compute most probable diagnoses, in combination with other measurement selection heuristics 
\cite{Rodler2013,rodler2018ruleML,rodler17dx_activelearning,moret1982decision}, or on diagnosis problems from other domains, such as spreadsheet \cite{jannach2014avoiding} or software debugging \cite{wotawa2010fault}, and \emph{(2)}~the integration of RBF-HS into our debugging tool \emph{OntoDebug}\footnote{See \url{http://isbi.aau.at/ontodebug}.} \cite{DBLP:conf/foiks/SchekotihinRS18}.
\vspace{10pt}

\noindent\textbf{Acknowledgments.}
This work was supported by the Austrian Science Fund (FWF), contract \mbox{P-32445-N38}.

\fontsize { 8.3pt }{ 8.8pt } 
\selectfont

\end{document}